\documentclass[10pt,twocolumn,letterpaper]{article}

\usepackage{cvpr}              
\usepackage{graphicx}
\usepackage{booktabs}
\usepackage{colortbl}
\usepackage[utf8]{inputenc} 
\usepackage{amsmath} 
\usepackage[table]{xcolor}
\usepackage[linesnumbered,ruled,vlined]{algorithm2e}
\usepackage{amsmath}
\usepackage{amssymb}
\newcommand{\rbf}[1]{\text{#1}} 

\usepackage[linesnumbered,ruled,vlined]{algorithm2e}
\usepackage{amsmath}
\usepackage{amsfonts}
\usepackage{xcolor}
\usepackage{bbding}
\usepackage{multirow}
\usepackage{tabularx}
\usepackage{makecell}
\usepackage[accsupp]{axessibility}  

\makeatletter
\newcommand\blfootnote[1]{%
  \begingroup
  \renewcommand\thefootnote{}\footnote{#1}%
  \addtocounter{footnote}{-1}%
  \endgroup
}
\makeatother

\SetKwComment{Comment}{\color{gray!80}// }{} 

\usepackage[dvipsnames]{xcolor} 
\usepackage{xparse}            
\usepackage[table]{xcolor}

\definecolor{R1color}{RGB}{188, 137, 53}  
\definecolor{R2color}{RGB}{95, 145, 62}   
\definecolor{R3color}{RGB}{101, 155, 175} 
\definecolor{HighlightPink}{RGB}{219, 48, 122} 

\definecolor{cvprblue}{rgb}{0.21,0.49,0.74}
\usepackage[pagebackref,breaklinks,colorlinks,allcolors=cvprblue]{hyperref}

\def\paperID{} 
\def\confName{CVPR}
\def\confYear{2026}

\title{CUBic: Coordinated Unified Bimanual Perception and Control Framework}

\author{
    Xingyu Wang$^{1,*}$ \quad
    Pengxiang Ding$^{2,3*,\dagger}$ \quad
    Jingkai Xu$^{4}$ \quad
    Donglin Wang$^{2}$ \quad
    Zhaoxin Fan$^{1,\text{\Envelope}}$ \\[1mm]
    $^{1}$Beijing Advanced Innovation Center for Future Blockchain and Privacy Computing, \\
    School of Artificial Intelligence, Beihang University \\
    $^{2}$Westlake University \quad $^{3}$Zhejiang University \quad
    $^{4}$Peking University \\[0.5mm]
    {\tt\small wangxingyu227@gmail.com} \quad
    {\tt\small dingpx2015@gmail.com} \quad
    {\tt\small zhaoxinf@buaa.edu.cn}
}

\begin{document}
\maketitle
\begin{abstract}
\blfootnote{$^*$ Equal contribution, $^\dagger$ Project lead, $^{\text{\Envelope}}$ Corresponding author.}
Recent advances in \textit{visuomotor policy learning} have enabled robots to perform control directly from visual inputs. 
Yet, extending such end-to-end learning from \textit{single-arm} to \textit{bimanual} manipulation remains challenging due to the need for both independent perception and coordinated interaction between arms. 
Existing methods typically favor one side—either decoupling the two arms to avoid interference or enforcing strong cross-arm coupling for coordination—thus lacking a unified treatment. 
We propose \textbf{CUBic}, a \textit{Coordinated and Unified framework for Bimanual perception and control} that reformulates bimanual coordination as a unified perceptual modeling problem. 
CUBic learns a shared tokenized representation bridging perception and control, where independence and coordination emerge intrinsically from structure rather than from hand-crafted coupling. 
Our approach integrates three components: unidirectional perception aggregation, bidirectional perception coordination through two codebooks with shared mapping, and a unified perception-to-control diffusion policy. 
Extensive experiments on the \textit{RoboTwin} benchmark show that CUBic consistently surpasses standard baselines, achieving marked improvements in coordination accuracy and task success rates over state-of-the-art visuomotor baselines.
\end{abstract}    
\section{Introduction}
\label{sec:intro}

\begin{figure*}[ht]
    \centering
    \includegraphics[width=1\linewidth]{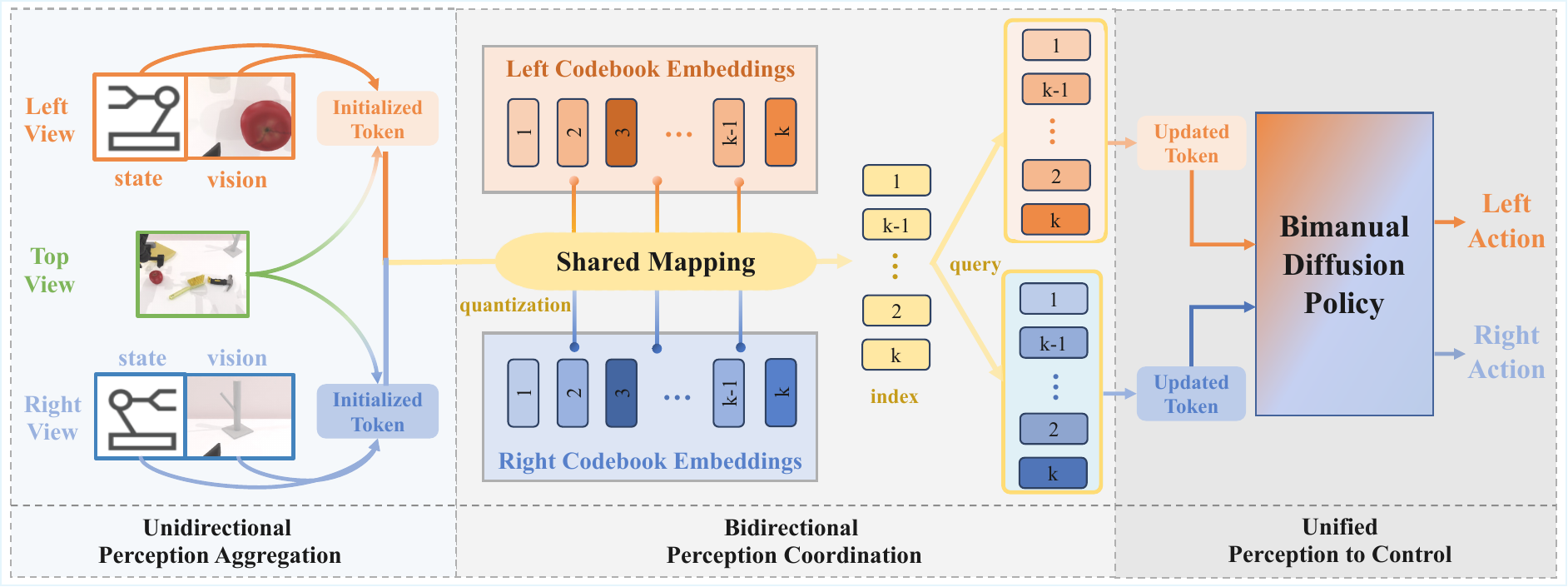}
    \caption{\textbf{Overall Framework.} Our proposed unified framework for bimanual perception and control collaboration. The unidirectional perception aggregation module leverages a unified masked attention mechanism to learn bimanual-relevant perceptual information from the top view; the bidirectional perception coordination module employs a dual-codebook shared mapping mechanism to achieve implicit coordination of bimanual perceptual information; the bimanual diffusion policy, built upon DiT, utilizes cross-attention to map bimanual perceptual information to actions. By isolating and integrating self-attention layers, we enable a two-stage unified training paradigm that seamlessly transitions from perceptual collaboration to action collaboration.}
    \label{fig:fig1}
\end{figure*}

Recent advances in \textit{visuomotor policy learning}  \cite{black2024pi0visionlanguageactionflowmodel, chi2024diffusionpolicy, fu2024mobile, liu2024rdt, gao2026vitavisiontoactionflowmatching, fu2025cordvipcorrespondencebasedvisuomotorpolicy} have significantly improved the ability of robots to perform control tasks directly from visual observations, enabling end-to-end mapping from images to actions. However, most existing works still focus on \textit{single-arm manipulation} \cite{chi2024diffusionpolicy, Ze2024DP3, 3d_diffuser_actor, noh20253dflowdiffusionpolicy, 11246340, Ma_2024_CVPR, Jia_2025_CVPR, zhang2024flowpolicyenablingfastrobust, zhu2023learninggeneralizablemanipulationpolicies}, and extending such visuomotor learning to \textit{coordinated bimanual manipulation} \cite{fu2024mobile, grotz2024peract2, liu2024rdt, lu2024anybimanual, xu2025diffusionbasedimaginativecoordinationbimanual, motoda2025learningbimanualmanipulationaction, lee2024interactinterdependencyawareaction, gkanatsios20253dflowmatchactorunified, hurova2025samplingbasedoptimizationparallelizedphysics} remains highly challenging.

The core difficulty in bimanual manipulation lies in learning \textbf{cross-arm coordination} \cite{grotz2024peract2, liu2024voxact, lu2024anybimanual, grannen2023stabilize, deng2025safebimanualdiffusionbasedtrajectoryoptimization} across both perception and control. Each arm must not only perceive and act independently, but also maintain \textbf{spatial and temporal consistency} \cite{tan2025anyposautomatedtaskagnosticactions, Lv_2025_CVPR, li2026momagengeneratingdemonstrationssoft, shen2025biassemblelearningcollaborativeaffordance, gao2025dagplangeneratingdirectedacyclic, yang2025gripperkeyposeobjectpointflow, Pan_2025_CVPR} with the other. Achieving such harmony demands cross-arm alignment of visual observations and motor actions—representing a level of coupling that visuomotor architectures struggle to model effectively.

To address this challenge, prior works have explored two primary directions. One line of research \cite{lu2024anybimanual, tan2025anyposautomatedtaskagnosticactions, motoda2025learningbimanualmanipulationaction, im2026twinvladataefficientbimanualmanipulation} seeks to \textit{decouple} the sensory inputs and control streams of each arm (e.g., \textit{AnyBimanual}), ensuring independence and reducing mutual interference. Another line \cite{grotz2024peract2, fu2024mobile, lee2024interactinterdependencyawareaction, 11271705} aims to \textit{enhance inter-arm communication} through mechanisms such as cross-attention, encouraging information exchange between arms. While both directions tackle valid subproblems, they inherently contradict each other: the former emphasizes separation, whereas the latter promotes interaction. This tension raises a fundamental question—\textbf{how can we unify these opposing objectives into a coherent framework for coordinated bimanual manipulation?}

We present \textbf{CUBic}—a \textit{Coordinated and Unified framework for Bimanual perception and control}—which reframes this contradiction through a shared, tokenized representation that naturally integrates independence and coordination. As illustrated in Fig.~\ref{fig:fig1}, CUBic consists of three key components. 
1) \textbf{Unidirectional Perception Aggregation} encodes multi-view sensory inputs (from left, right, and top views) into arm-specific perceptual tokens that capture localized visual and proprioceptive cues. 
2) \textbf{Bidirectional Perception Coordination} establishes a shared codebook space, allowing each arm to query and update the other's token embeddings—achieving both autonomy and mutual awareness within a unified perceptual structure. 
3) \textbf{Unified Perception-to-Control Diffusion Policy} consumes the coordinated tokens to generate synchronized, physically consistent action trajectories for the two arms.
Unlike previous approaches that alternately emphasize either inter-arm independence \cite{grannen2023stabilize, liu2024voxact, lu2024anybimanual} or interaction \cite{grotz2024peract2, fu2024mobile, lee2024interactinterdependencyawareaction}, CUBic reconceptualizes bimanual coordination as a unified perceptual modeling problem.
By learning a shared, tokenized representation that bridges perception and control, CUBic allows coordination to emerge intrinsically from the model’s structure rather than being enforced through explicit coupling mechanisms.

Modeling both separation and interaction as emergent properties of a shared token space, CUBic closes the gap between perception and control, enabling coherent bimanual reasoning and physically consistent execution within a single unified framework.
Extensive experiments on the \textit{RoboTwin} benchmark \cite{Mu_2025_CVPR} demonstrate that CUBic consistently outperforms conventional Diffusion Policy (DP) \cite{chi2024diffusionpolicy} and even DP3 \cite{Ze2024DP3} \textbf{without any 3D perception}, yielding substantial improvements in coordination accuracy and task success rates over state-of-the-art visuomotor policies.

\section{Related Work}
\label{sec:related_work}

\subsection{Visuomotor Policy for Bimanual Manipulation}

End-to-end visuomotor policy learning has achieved impressive results in enabling robots to perform manipulation directly from raw visual observations~\cite{chi2024diffusionpolicy, fu2024mobile, liu2024rdt}. 
Most of these methods, however, are designed for single-arm scenarios \cite{chi2024diffusionpolicy, Ze2024DP3, xu2025speci, su2025dspv2improveddensepolicy, tian2026vitasvisualtactilesoft, Su_2025_ICCV}, where perception and control are modeled for an individual manipulator. 
As a result, they struggle to generalize to multi-arm or cooperative settings, where inter-arm dependencies and spatio-temporal coordination become essential.
Compared with single-arm tasks, bimanual collaborative manipulation introduces substantially greater complexity \cite{grannen2023stabilize, gu2023maniskill2, Mu_2025_CVPR, zhou2025you, gao2024bi, wu2025robocoinopensourcedbimanualrobotic}. 
The joint action space expands combinatorially with the degrees of freedom of two arms, and coordination requires reasoning over multiple spatial and temporal constraints~\cite{xie2020deep, franzese2023interactive}. 
This high-dimensional coupling often challenges the scalability of end-to-end imitation learning frameworks~\cite{fu2024mobile, liu2024rdt, black2024pi0visionlanguageactionflowmodel, im2026twinvladataefficientbimanualmanipulation}. 
To cope with this difficulty, previous works have incorporated inductive biases to simplify the learning space—for example, assigning stability and functionality roles to the two arms~\cite{grannen2023stabilize}, parameterizing motion primitives~\cite{chitnis2020efficient, franzese2023interactive}, or adopting voxelized spatial representations~\cite{grotz2024peract2, liu2024voxact, lu2024anybimanual}. 
Although such strategies improve training efficiency, they largely rely on task-specific design assumptions and are limited in expressiveness for diverse and multimodal cooperative behaviors.

More broadly, existing frameworks for bimanual manipulation can be divided into two major paradigms. 
The first decouples perception and control for each arm to minimize mutual interference~\cite{lu2024anybimanual, grannen2023stabilize, im2026twinvladataefficientbimanualmanipulation}, ensuring stability and modularity but sacrificing cross-arm consistency. 
The second explicitly promotes inter-arm communication through attention-based or message-passing mechanisms~\cite{fu2024mobile, grotz2024peract2, lee2024interactinterdependencyawareaction}, improving coordination but often at the cost of disentanglability and robustness. 
Both directions address important aspects of bimanual manipulation, yet their objectives—\textit{independence} versus \textit{interaction}—are inherently conflicting. 
In contrast, our approach reformulates bimanual coordination as a unified representation problem rather than a structural trade-off. 
By modeling intrinsic semantic coordination in the latent space, our method achieves coherent dual-arm reasoning and flexible manipulation behaviors without enforcing explicit role partitioning \cite{grannen2023stabilize} or relying on handcrafted structural constraints \cite{lu2024anybimanual}.

\subsection{Unified Representation through Tokenization}
Achieving a unified perceptual–action representation is central to bridging perception and control in visuomotor policy learning. 
Most existing robot learning frameworks treat perception and action as separate modules \cite{Ze2024DP3, chi2024diffusionpolicy}, limiting their ability to reason jointly about sensory context and multi-arm coordination. 
Recent advances in vector-quantized (VQ) tokenization have shown strong potential for representing both visual~\cite{qu2025tokenflow, lu2025h3dptriplyhierarchicaldiffusionpolicy} and action~\cite{bu2025univla, cui2024dynamo, ye2024latent, schmidt2023learning, Wang_2025_ICCV, Chen_2025_ICCV, chen2025villa} modalities in a compact and structured manner. 
By discretizing continuous latent spaces into codebook tokens, these methods enable efficient and compositional learning across vision and control.
Some approaches adopt multi-scale VQ schemes to encode semantic and pixel-level visual cues~\cite{qu2025tokenflow}, thereby supporting coarse-to-fine action generation~\cite{lu2025h3dptriplyhierarchicaldiffusionpolicy, gong2025carpvisuomotorpolicylearning}. 
Others utilize unsupervised VQ-based frameworks to extract latent action tokens from videos, improving cross-embodiment and cross-task generalization~\cite{bu2025univla, ye2024latent, chen2025villa, Chen_2025_ICCV}. 
Despite their success, these methods are typically developed for single-arm settings and cannot directly model inter-arm dependencies or perceptual alignment across manipulators.

Building on these insights, our framework leverages tokenization as the foundation of a unified representation for bimanual perception and control. 
Specifically, we introduce two VQ codebooks with shared mapping that encodes coordinated perceptual and action constraints into a common latent token space, allowing information to be exchanged naturally between arms. 
This shared representation serves as the structural bridge between perception and control, enabling end-to-end learning of synchronized bimanual manipulation policies.

\section{Preliminaries}

\begin{figure*}[ht]
    \centering
    \includegraphics[width=0.9\linewidth, height=7.5cm]{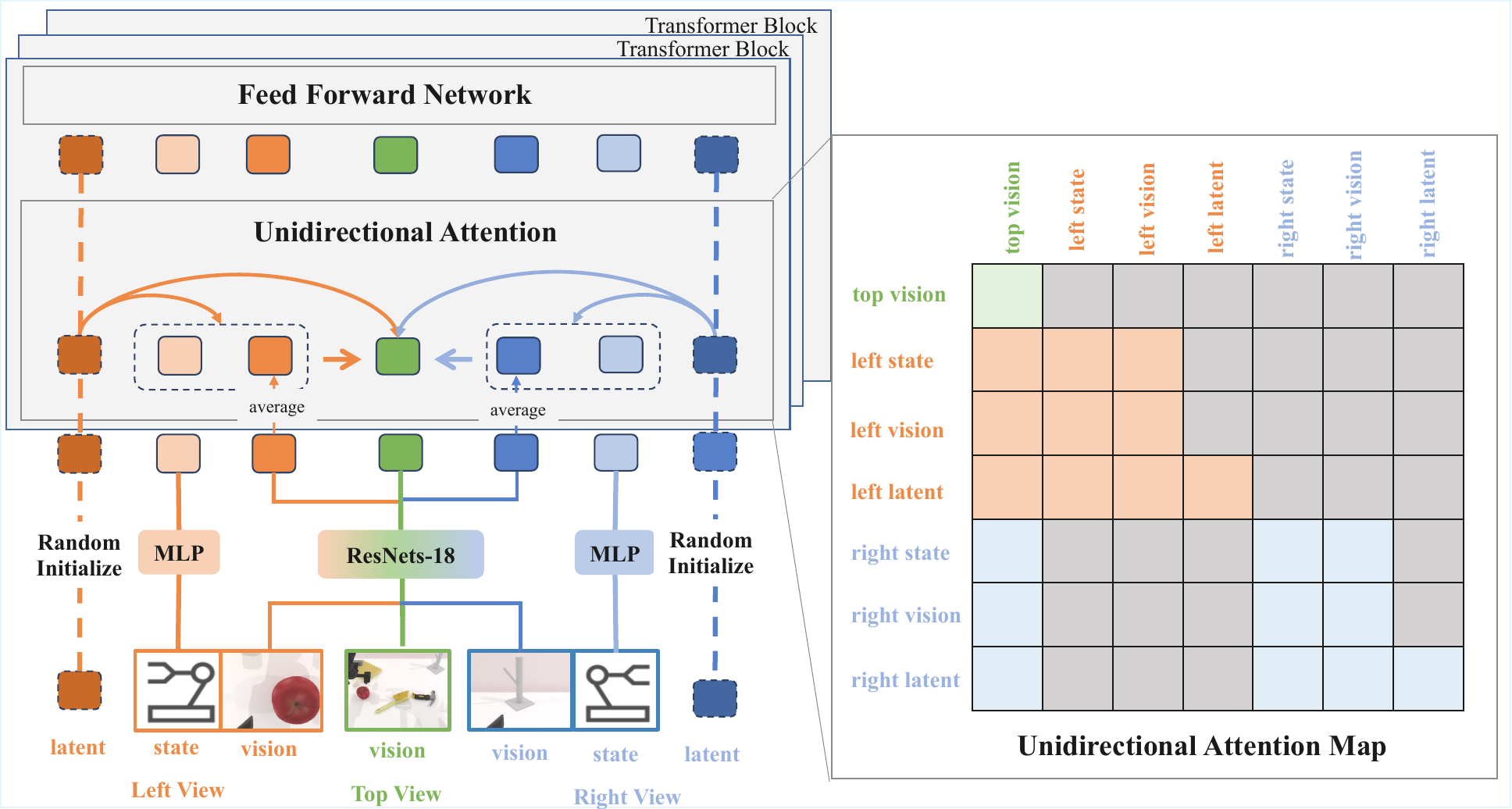}
    \caption{\textbf{Unidirectional Perception Aggregation.} Multi-view images are encoded using independent ResNet-18 backbones, while joint information is projected to the same dimensionality via MLPs. To establish an implicit bridge for perceptual collaboration, we introduce learnable latent tokens initialized randomly. In the unidirectional attention mechanism, a masking strategy ensures that each arm learns relevant information from the top view exclusively through its own wrist camera and joint states; conversely, the top view attends only to itself, thereby preventing information leakage.}
    \label{fig:fig2}
\end{figure*}

\subsection{Problem Formulation}

Problem formulation will consider a task $\mathcal{T}$, where there are $N$ expert demonstrations $\{\tau_i\}_{i=1}^N$. Each demonstration $\tau_i$ is a sequence of state-action pairs. We formulate robot imitation learning as an action sequence prediction problem, training a model to minimize the error in future actions conditioned on historical states. Specifically, imitation learning minimize the behavior cloning loss $\mathcal{L}_{bc}$ formulated as
\begin{equation}
\mathcal{L}_{bc} = \mathbb{E}_{s, a \sim \mathcal{T}} \left[ \sum_{t=0}^T \mathcal{L}(\pi_\theta(a_H|s_O), a_H) \right],
\end{equation}
where $a$ represents the action, $s$ denotes the state or observation according to the specific task description, $t$ is the current time step, $H$ is the prediction horizon, and $O$ is the historical horizon. For notational simplicity, we denote the action sequence $a_{t:t+H-1}$ as $a_H$ and the state sequence $s_{t-O+1:t}$ as $s_O$. Here, $\mathcal{L}_{bc}$ represents a supervised action prediction loss, such as mean squared error or negative log-likelihood, $T$ is the length of the demonstration, and $\theta$ represents the learnable parameters of the policy network $\pi_\theta$.

\subsection{Diffusion Policy}
 Diffusion-based policies \cite{chi2024diffusionpolicy} utilize Denoising Diffusion Probabilistic Models \cite{ho2020denoising} to approximate the conditional distribution $p(a_H|s_O)$. The core idea of diffusion models is to continuously add Gaussian noise to make action sequences a Gaussian and leverage the denoising process for generating actions. Let $a_H^0$ represent a real action sample, the forward process aims to add Gaussian noise and result in a set of noisy data, i.e., $q(a_H^t|a_H^{t-1}) = \mathcal{N}(a_H^t; \sqrt{\alpha_t}a_H^{t-1}, (1-\alpha_t)\mathbf{I})$, where $a_H^t$ and $\alpha_t$ indicate the noisy data and noise amplitude at the timestep $t$. Let $\bar{\alpha}_t = \prod_{i=1}^t \alpha_i$, the above process can be simplified as:

\begin{equation}
a_H^t = \sqrt{\bar{\alpha}_t}a_H^0 + \sqrt{1 - \bar{\alpha}_t}\epsilon_t
\end{equation}

The reverse process starts from the most noisy sample $a_H^T$ can be described in a variational approximation of the probabilities $q(a_H^{t-1}|a_H^t)$, as follows:
\begin{equation}
p(a_H^{t-1}|a_H^t) = \mathcal{N}(a_H^{t-1}; \sqrt{\bar{\alpha}_{t-1}}\mu_\theta(a_H^t, t), (1 - \bar{\alpha}_{t-1})\mathbf{I})  \notag
\end{equation}
where $\mu_\theta(a_H^t, t) = (a_H^t - \sqrt{1 - \bar{\alpha}_t}\epsilon_\theta(a_H^t, t)) / \sqrt{\bar{\alpha}_t}$ is a learnable neural network to estimate $a_H^{t-1}$. Further, in diffusion policy, the denoising process learns the noise estimator $\epsilon_\theta(a_H^t, s_O)$ to controll the action generation based on the current observation.

\section{Method}

Our proposed \textbf{CUBic} framework aims to achieve intrinsically coordinated bimanual manipulation by learning a unified perceptual representation that bridges perception and control. 
The framework integrates \textbf{unidirectional perception aggregation} to fuse multi-view sensory observations into arm-specific perceptual tokens, \textbf{bidirectional perception coordination} to establish mutual perceptual awareness within a shared codebook space while preserving arm-level autonomy, and a \textbf{unified perception-to-control} diffusion policy that transforms the coordinated tokens into synchronized, physically consistent dual-arm trajectories. 
Together, these components form a unified perception–control pathway in a shared tokenized latent space, allowing coordination to emerge naturally from the model’s architecture rather than through explicit coupling mechanisms.
Below, we elaborate on these three components in sequence.

\subsection{Unidirectional Perception Aggregation}

\noindent\textbf{Insight.}  
In bimanual manipulation with multi-view inputs, each camera view contributes complementary visual information. 
Head-mounted or external views provide coarse global context, capturing both arms and their relative positions, while wrist-mounted cameras deliver fine-grained local observations, such as object–gripper relationships and surface details, combined with joint state information of each manipulator. 
Directly fusing all these modalities often complicates training, dilutes the utility of local–global cues, and introduces redundant correlations. 
To mitigate this, we categorize the perceptual inputs: wrist-camera feeds and joint states are treated as arm-specific information for local perception, whereas head-mounted or external views serve as a shared global context for cross-arm collaboration. 
Accordingly, we employ a masked-attention transformer for unidirectional perception aggregation.

\noindent\textbf{Details.}  
As illustrated in Fig.~\ref{fig:fig2}, semantic features are extracted from each RGB view using a ResNet-18 backbone, and proprioceptive signals are projected into the same feature space via a lightweight MLP. 
To initialize latent bimanual coordination, we define two sets of learnable latent action tokens, $a_q^\text{left} \in \mathbb{R}^{N \times d}$ and $a_q^\text{right} \in \mathbb{R}^{N \times d}$, where $N$ denotes the number of latent tokens, representing the latent action spaces of the left and right arms, respectively. 
A transformer equipped with unidirectional self-attention is then utilized to model the latent action representations, where an attention mask enforces information decoupling between the two arms. 
For the left arm, the arm-specific token sequence $c_\text{left}$ is formed by concatenating the globally averaged wrist-camera features with its proprioceptive embedding. 
The latent tokens $a_q^\text{left}$ attend only to their corresponding arm-specific tokens and the shared head-camera features, while the arm-specific tokens attend exclusively to the shared features. 
The right-arm branch follows the same symmetric design.

\noindent\textbf{Analysis.}  
This configuration serves two primary purposes: it enforces strict information insulation between the two arms, preventing premature cross-arm interference, and it anchors both latent action spaces to a common perceptual foundation derived from the shared head view. 
By disentangling arm-specific cues while maintaining a unified global context, the model achieves stable and interpretable perceptual representations that balance local precision and global awareness. 
Consequently, both manipulators reason over consistent contextual information, forming a coherent latent structure that naturally lends itself to the subsequent bidirectional perception coordination stage.

\begin{figure*}[ht]
    \centering
    \includegraphics[width=1\linewidth, height=7cm]{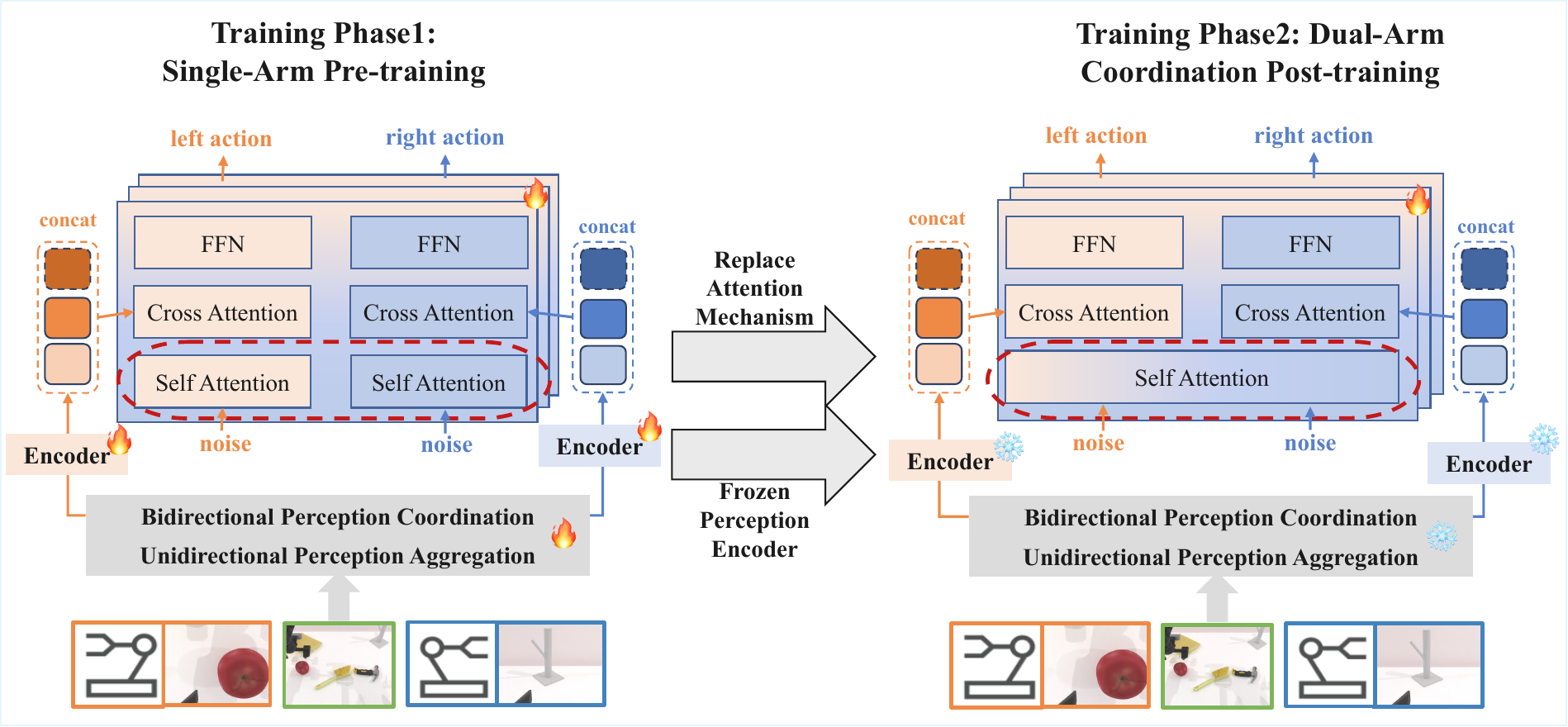}
    \caption{\textbf{Training Recipe.} We employ a two-stage training paradigm to progressively cultivate collaborative perception and control. In the first stage, actions for each arm are decoded independently, enabling the model to acquire collaborative strategies at the perception level. In the second stage, the decoding processes for both arms are fused while the perception module is frozen, thereby preserving the pre-trained perceptual collaboration capabilities and facilitating the learning of action collaboration at the control level. This progressive approach ultimately yields a unified framework for perception and action collaboration.}
    \label{fig:fig3}
\end{figure*}

\subsection{Bidirectional Perception Coordination}

\noindent\textbf{Insights.}  
This module establishes coordinated relationships between the latent actions of both arms, enabling controlled decoupling and inter-arm awareness. 
After the masked self-attention stage, the latent action tokens $a_q^\text{left}$ and $a_q^\text{right}$ each encode implicit information derived from the shared head-camera view. 
These tokens are then quantized using two dual codebooks: 
$Z_{\text{left}} = \{a^\text{left}_{z,i}\}_{i=1}^K \in \mathbb{R}^{K \times d_z}$ and 
$Z_{\text{right}} = \{a^\text{right}_{z,i}\}_{i=1}^K \in \mathbb{R}^{K \times d_z}$, 
where $K$ denotes the number of codebook entries.  
Both codebooks share a unified mapping space, allowing each arm’s latent representations to be quantized while jointly considering their mutual context.

\noindent\textbf{Details.}  
Given the encoded latent tokens $a_q^\text{left}$ and $a_q^\text{right}$, their respective distances to the codebook entries $d_{\text{left},i}$ and $d_{\text{right},i}$ are computed as:
\begin{align}
    d_{\text{left},i} &= \|a_q^\text{left} - a^\text{left}_{z,i}\|_2^2, \text{ for } i = 1, \dots, K \\
    d_{\text{right},i} &= \|a_q^\text{right} - a^\text{right}_{z,i}\|_2^2, \text{ for } i = 1, \dots, K\\
    i^* &= \arg \min_i (d_{\text{left},i} + d_{\text{right},i})
\end{align}
The optimal quantization index $i^*$ minimizes the combined distance, encouraging convergence toward a latent representation jointly consistent across arms. 
To further enhance representational capacity, we apply a residual vector quantization (RVQ) hierarchy~\cite{lee2022autoregressive}, which improves codebook expressiveness and stabilizes training.  

\noindent\textbf{Analysis.}  
Through this shared quantization process, the dual codebooks learn a joint distribution of left- and right-arm latent features, thereby establishing an intrinsic coupling mechanism between otherwise decoupled perceptual streams. 
This coupling not only aligns the semantic representations of both arms within a common latent manifold, but also preserves their functional independence, allowing cross-arm correlations to emerge implicitly through shared codebook dynamics. 
As a result, the model attains a balanced representation that supports both arm-specific reasoning and coordinated joint behavior, providing a unified perceptual foundation for the subsequent diffusion-based control stage.

\subsection{Unified Perception to Control}

\noindent\textbf{Insights.} 
After obtaining perceptual features that balance collaborative context and arm-specific precision, the next step is to translate these representations into executable actions. 
Unlike conventional DiT-based policies, we adopt a two-stage training paradigm to progressively acquire dual-arm control capabilities:  
(1) \textbf{Single-Arm Pre-training}, where each arm is trained independently with a transformer composed of self-attention, cross-attention, and feed-forward layers. 
This stage focuses solely on learning the interaction between perceptual embeddings and action tokens, enabling the model to master individual control skills and ensuring stable convergence during discrete representation learning.  
(2) \textbf{Dual-Arm Coordination Post-training}, where the model transitions from independent control to coordinated bimanual operation. 
All perception-encoding parameters are frozen, and the DiT architecture is modified so that the two arms share self-attention layers, allowing mutual visibility and introducing structured correlations in the diffusion noise. 
Meanwhile, cross-attention connections to perceptual tokens remain separated, preserving the integrity of perception-to-control flow. 
This hierarchical design maintains perceptual disentanglement while fostering synchronized and physically consistent dual-arm actions.

\noindent\textbf{Details.}  
The quantized latent action tokens $a_z^\text{left}$ and $a_z^\text{right}$ are first fed into an additional encoder module, along with their corresponding post-attention arm-specific tokens $c_\text{left}$ ($c_\text{right}$) and the head-camera tokens. 
This process transforms the discrete latent action information into enriched visual-semantic embeddings, denoted as $Q_\text{left}$ and $Q_\text{right}$. 
These embeddings provide the perceptual conditions for subsequent action decoding:
\begin{align}
    Q_\text{left} &= \text{concat}(a_z^\text{left}, c_\text{left}), \\
    Q_\text{right} &= \text{concat}(a_z^\text{right}, c_\text{right}).
\end{align}

For action decoding, we employ a Diffusion Transformer (DiT)~\cite{Peebles2022DiT} as the policy network to generate action sequences $a_H \in A$ conditioned on $Q$. 
The perceptual embeddings $Q_\text{left}$ and $Q_\text{right}$ are integrated into the diffusion transformer blocks through cross-attention, enabling the diffusion process to exploit both arm-specific and shared perceptual cues. 
The policy aims to reconstruct the original clean action trajectory $a_H^0$ from its progressively noised version:
\begin{equation}
    a_H^k = \sqrt{\bar{\alpha}_k}\,a_H^0 + \sqrt{1 - \bar{\alpha}_k}\,\epsilon,
\end{equation}
where $\epsilon$ represents Gaussian noise and $\bar{\alpha}_k$ denotes the variance schedule coefficient at step $k$. 
Accordingly, the model learns a denoising function $D_\theta(a_H^k, k, Q)$ optimized under the standard diffusion objective:
\begin{equation}
    \mathcal{L}_{\text{diff}}(\theta; A) =
    \mathbb{E}_{a_H^0, \epsilon, k}\big[\| D_\theta(a_H^k, k, Q) - \epsilon \|^2\big].
\end{equation}
We adopt $k = 100$ diffusion steps during training with a cosine noise schedule, and use a deterministic DDIM~\cite{song2022denoisingdiffusionimplicitmodels} sampling strategy for inference, reducing the required steps to $k = 10$.

\begin{figure*}[ht]
    \centering
    \includegraphics[width=0.7\textwidth, height=6.8cm]{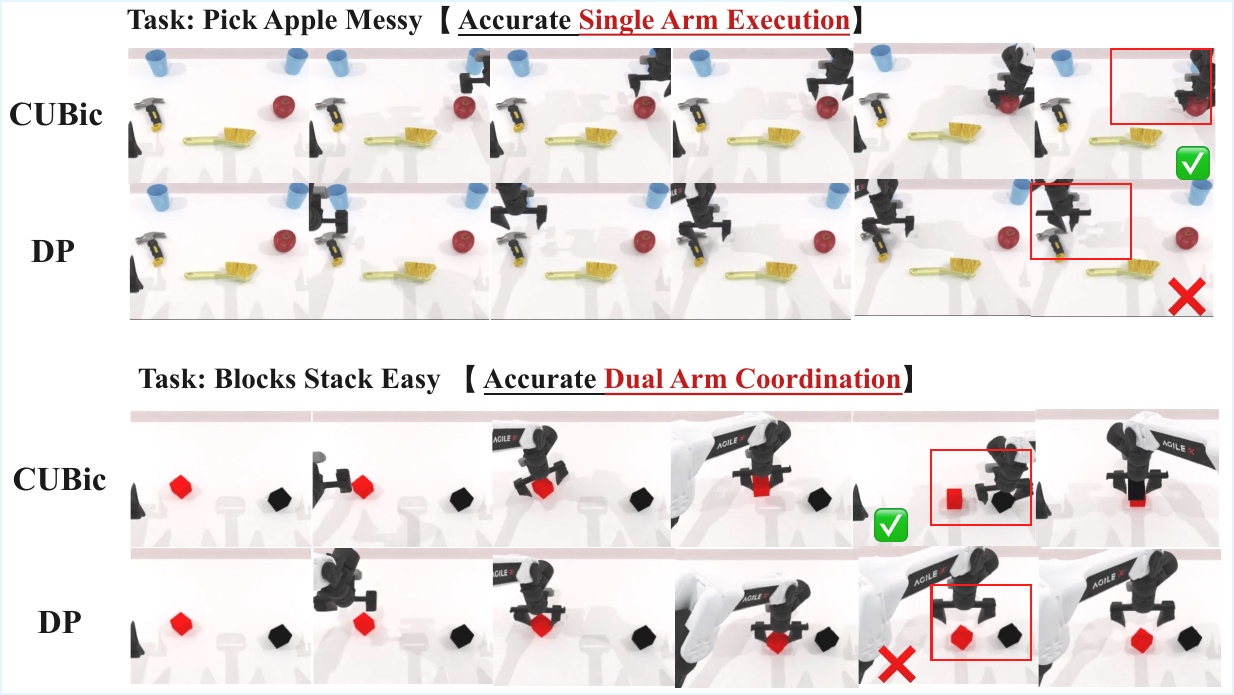}
    \caption{\textbf{Visualization in RoboTwin.} 
    As illustrated, compared with Diffusion Policy (DP), CUBic demonstrates superior performance in object localization and precise single-arm grasping, while maintaining strong coordination in bimanual manipulation scenarios.}
    \label{fig:fig4}
\end{figure*}

\noindent\textbf{Training Recipe.}  
We adopt a two-stage training strategy, as shown in Figure~\ref{fig:fig4}, to progressively build dual-arm coordination capability from perception to control.

\noindent(1)~\textit{Pre-training stage.}  
In the first stage, the two policy branches are trained independently to stabilize learning and strengthen the perception encoder’s ability to produce consistent representations. 
Each arm employs an independent transformer with its own self-attention, cross-attention, and feed-forward modules, guaranteeing that the perception branch learns disentangled but informative cues. We evenly partition $a_H$ into two components, $a_{H,\text{left}}$ and $a_{H,\text{right}}$, along the direction of the motion dimension.
The diffusion losses for the two arms are defined as:
\begin{align}
    \mathcal{L}_{\text{diff}}^\text{left}(\theta; A) &= 
    \mathbb{E}_{a_{H,\text{left}}^0, \epsilon, k}\| D_\theta(a_{H,\text{left}}^k, k, Q_\text{left}) - \epsilon \|^2, \\
    \mathcal{L}_{\text{diff}}^\text{right}(\theta; A) &= 
    \mathbb{E}_{a_{H,\text{right}}^0, \epsilon, k}\| D_\theta(a_{H,\text{right}}^k, k, Q_\text{right}) - \epsilon \|^2.
\end{align}

Vector quantization is trained with a straight-through gradient estimator 
$a_z = \text{sg}[a_z - a_q] + a_q$, 
where $\text{sg}[\cdot]$ denotes the stop-gradient operator. 
The codebook objective is formulated as:
\begin{equation}
    \mathcal{L}_{\text{VQ}} =
    \|\text{sg}[a_q] - a_z\|_2^2 + \beta\|a_q - \text{sg}[a_z]\|_2^2,
\end{equation}
where the latter term represents the commitment loss weighted by $\beta$. 
The overall pre-training objective is then:
\begin{equation}
    \mathcal{L}_{\text{phase1}} =
    \mathcal{L}_{\text{diff}}^\text{left} + \mathcal{L}_{\text{VQ}} + \mathcal{L}_{\text{diff}}^\text{right}.
\end{equation}

\noindent(2)~\textit{Post-training stage.}  
In the second stage, we freeze all perception modules to preserve their learned collaborative representations, and merge the self-attention layers within the diffusion transformer into a unified module visible to both arms. 
This modification introduces full interaction between the two arms at the policy level, enabling explicit cross-arm coordination while maintaining perceptual independence through separate cross-attention pathways. 
The total post-training objective is expressed as:
\begin{equation}
    \mathcal{L}_{\text{phase2}} =
    \mathcal{L}_{\text{diff}}^\text{left} + \mathcal{L}_{\text{diff}}^\text{right}.
\end{equation}

Through this two-phase pipeline, the model gradually transitions from learning perceptual cooperation to executing coordinated bimanual control, resulting in a unified and interpretable perception-to-control mapping.

\section{Experiment}

\begin{table*}[ht]
    \centering
    \resizebox{0.98\linewidth}{!}{
    \begin{tabular}{@{}l l l l l  l l l l l @{}}
        \toprule
         & \cellcolor{gray!20} Avg.$\uparrow$ & Pick Apple & Dual Bottles & Blocks & Dual Bottles & Block & Dual Shoes  & Put Apple  \\
        Method &\cellcolor{gray!20} Success $(\%)$  & Messy & Pick (Easy) & Stack (Easy) & Pick (Hard) & Handover & Place & Cabinet  \\
        \midrule
        DP3~\cite{Ze2024DP3} & \cellcolor{gray!20} 39.8 & 9.7 \rbf{± 2.1} & 55.3 \rbf{± 11.5} & - & \underline{58.0} \rbf{± 3.0} & \underline{77.3} \rbf{± 11.6}  & \textbf{12.0} \rbf{± 1.7}  & \underline{66.3} \rbf{± 22.3} \\
        DP~\cite{chi2024diffusionpolicy} & \cellcolor{gray!20} 38.5 & \underline{29.3} \rbf{± 5.0} & \textbf{85.7} \rbf{± 6.7} & \underline{8.0} \rbf{± 4.4} & \textbf{59.3} \rbf{± 5.5} & 76.0 \rbf{± 16.1} & 3.0 \rbf{± 1.0} & 8.0 \rbf{± 12.2} \\
        GR-MG \cite{li2024grmgleveragingpartiallyannotated} & \cellcolor{gray!20} 8.00 &  8.0 \rbf{± 7.9} &30.3 \rbf{± 7.1} & 0.0 \rbf{± 0.0} &17.7 \rbf{± 7.7} & 0.0 \rbf{± 0.0} &	0.0 \rbf{± 0.0}  & 0.0 \rbf{± 0.0}\\
        Ours & \cellcolor{gray!20} \textbf{51.8} & \textbf{40.0} \rbf{± 5.0} & \underline{84.3} \rbf{± 2.3} & \textbf{16.0} \rbf{± 0.0} & \underline{58.0} \rbf{± 2.0} & \textbf{85.7} \rbf{± 9.1} & \underline{10.0} \rbf{± 2.5} & \textbf{68.7} \rbf{± 4.6} \\
        \bottomrule
    \end{tabular}
    }
    \caption{\textbf{Evaluation on RoboTwin benchmark.} We report the mean and standard deviation of success rates averaged over 3 random seeds.  Best score in \textbf{bold}, second-best \underline{underlined}. Our method outperforms the previous state-of-the-art with an average elevation of 12.0\% success rate across 7 tasks. - denotes that color-specific tasks are not suitable for DP3 while only taking point cloud as input.} 
    \label{tab: eval}
    \vspace{-3mm}
\end{table*}

In Sections \ref{subsec:Comparison with the State-of-the-Art Methods} and \ref{subsec:Real-World Experiment}, we evaluate the manipulation capability of our proposed CUBic by presenting the experimental settings and results from both simulation \cite{Mu_2025_CVPR} and real-world tasks, respectively. The effectiveness of each component is validated through an ablation study in Section \ref{subsec:Ablation Study}. 

\subsection{Simulation Experiment}
\label{subsec:Comparison with the State-of-the-Art Methods}

\noindent\textbf{Benchmark.}  
We use the RoboTwin \cite{Mu_2025_CVPR} benchmark suite, which is widely adopted for evaluating bimanual tasks \cite{xu2025diffusionbasedimaginativecoordinationbimanual,bi2025h,lu2025h3dptriplyhierarchicaldiffusionpolicy}. RoboTwin benchmark is built on ManiSkill \cite{gu2023maniskill2} and features more complex scenarios. Expert demonstrations are generated via 3D generative models and LLMs, enabling diverse and realistic task variations. Each task includes 100 demonstrations, spanning 250 to 850 steps.

\noindent\textbf{Baselines.}  We compare our method against several baselines to validate the effectiveness:
\begin{itemize}
    
    \item \textbf{Diffusion Policy} \cite{chi2024diffusionpolicy}: a visuomotor policy learning framework that formulates action prediction as a conditional denoising diffusion process.
    \item \textbf{3D Diffusion Policy} \cite{Ze2024DP3}: leverages a lightweight MLP encoder to process sparse point clouds and a conditional diffusion model to generate actions, enabling efficient and robust visuomotor policy learning.
    \item \textbf{GR-MG} \cite{li2024grmgleveragingpartiallyannotated}: an enhanced version of GR-1 \cite{wu2023unleashinglargescalevideogenerative}, retaining the same model architecture and next-frame prediction loss. It improves GR-1 \cite{wu2023unleashinglargescalevideogenerative} by incorporating a multimodal, goal-conditioned generative policy that integrates language and goal images for robotic manipulation. 
\end{itemize}

\noindent\textbf{Metrics.} For each task, we report the mean and standard deviation of the success rate over three random initial seeds. The reuslt for each seed is evaluated across 50 executions. 

\noindent\textbf{Implementation Details.} For baseline models, we follow the same implementation and training configurations provided by \cite{chi2024diffusionpolicy} \cite{Ze2024DP3}
. For CUBic, we set the observation horizon $O=1$ and the prediction horizon $H=8$. We set the number of learnable latent tokens to $N=4$. For the VQ module, we use a codebook capacity of $K=256$ and a latent dimension of $d=32$.The model is trained on 4 NVIDIA 4090 GPUs with a per-GPU batch size of 32, resulting in a total batch size of 128. The training process includes two stages, and we train for 900 epochs in each stage. For evaluation, we use the latest saved checkpoint.


\noindent\textbf{Quantitative Results.}
As shown in Table~\ref{tab: eval}, our method achieves an average success rate of 51.8\% across seven bimanual manipulation tasks, surpassing DP3 by 12.0\%, DP by 13.3\%, and GR-MG by 43.8\%.
Notably, CUBic shows pronounced advantages in complex, long-horizon coordination tasks such as ``Pick Apple Messy'' and ``Blocks Stack Easy,'' indicating that the proposed unified codebook coordination mechanism substantially enhances bimanual manipulation capability.

\noindent\textbf{Qualitative Results.}
As illustrated in Figure~\ref{fig:fig4}, compared with DP, CUBic exhibits superior object localization, more precise single-arm grasping, and stronger inter-arm coordination in bimanual manipulation scenarios. 
This result also demonstrates the benefit of the shared codebook design, which maintains independent environmental perception for each arm while promoting synchronized collaboration between both arms.

\subsection{Real-World Experiment}
\label{subsec:Real-World Experiment}

\noindent\textbf{Experiment Setup.} Our real-world setup features a dual-arm Agibot with three Intel RealSense D435 cameras mounted on the head and both wrists. To comprehensively evaluate manipulation capabilities, we design six tasks with varying difficulty, target objects, and action sequences: (1) Apple to Plate, (2) Bowl to Plate, (3) Dual Banana Grasp, (4) Dual Bottle Grasp, (5) Banana Handover to Plate, and (6) Apple to Drawer.

\noindent\textbf{Data and Metrics.} For each task, we collect 100 expert trajectories and conduct 50 evaluation trials in varying spatial setups. We introduce a step-wise scoring system \ref{table:real-world setup}, where the total execution scores is the sum of points awarded for each successfully completed sub-task. We compare our approach against Diffusion Policy \cite{chi2024diffusionpolicy} as the baseline.

\begin{table}[ht]
\centering
\small
\setlength{\tabcolsep}{4pt}
\renewcommand{\arraystretch}{1.1}
\captionsetup{font=footnotesize}
\resizebox{\columnwidth}{!}{
\begin{tabular}{l|ccccc}
\toprule
& \textbf{Step1} & \textbf{Step2} & \textbf{Step3} & \textbf{Step4} & \textbf{Total Score} \\
\midrule
1 & Pick(50) & Place(50) & -- & -- & 100 \\
2 & Pick(50) & Place(50) & -- & -- & 100 \\
3 & L/R-Grasp(25/25) & L/R-Grasp(25/25) & -- & -- & 100 \\
4 & L/R-Grasp(25/25) & L/R-Grasp(25/25) & -- & -- & 100 \\
5 & Grasp(30) & Handover(40) & Place(30) & -- & 100 \\
6 & Grasp Apple(25) & Open Drawer(25) & Place Apple(25) & Close Drawer(25) & 100 \\
\noalign{\vspace{8pt}}
\multicolumn{6}{c}{
\includegraphics[width=1.5\columnwidth]{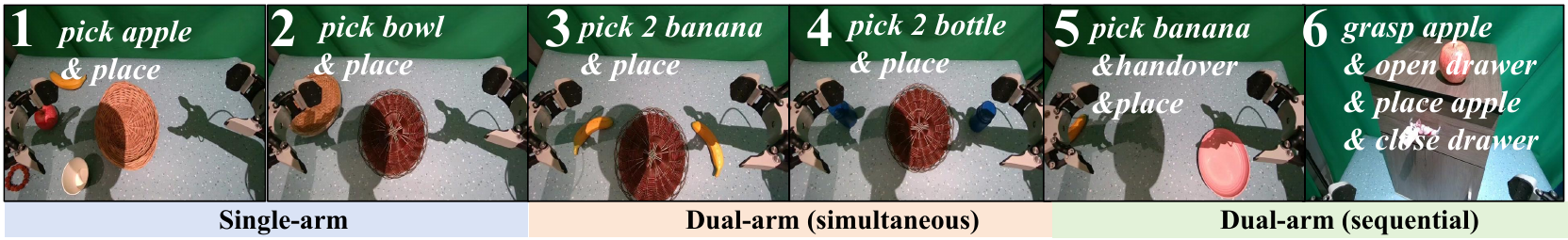}
} \\
\end{tabular}
}
\caption{\textbf{Real Task Definitions and Rules.}
(L/R means Left/Right)}
\label{table:real-world setup}
\end{table}

\noindent\textbf{Result.}
We evaluate models on both seen object positions from the training set (In-Domain) and randomly placed positions (Out-of-Domain). The table shows our method generally outperforms DP, and changes in object location do not significantly affect the final results.

\begin{table}[ht]
    \centering
    \small
    \renewcommand{\arraystretch}{1.0}
    \setlength{\tabcolsep}{6pt} 

    \resizebox{\linewidth}{!}{%
    \begin{tabular}{l cc cc}
        \toprule
        \multirow{2}{*}{\textbf{Task}} & \multicolumn{2}{c}{\textbf{In-Domain}} & \multicolumn{2}{c}{\textbf{Out-of-Domain}} \\
        \cmidrule(lr){2-3} \cmidrule(lr){4-5}
        & DP & \textbf{CUBic} & DP & \textbf{CUBic} \\
        \midrule
        Apple to Plate   & 45.0 & \textbf{50.0} & 25.0 & \textbf{35.0} \\
        Bowl to Plate    & 20.0 & \textbf{37.5} & 10.0 & \textbf{35.0} \\
        Banana Grasp     & 15.0 & \textbf{36.3} & 12.5 & \textbf{31.3} \\
        Bottle Grasp     & 10.0 & \textbf{77.5} & 7.5  & \textbf{57.5} \\
        Banana Handover  & 3.0  & \textbf{18.5} & 6.0  & \textbf{17.0} \\
        Apple to Drawer  & 23.8 & \textbf{38.8} & 15.0 & \textbf{32.5} \\
        \midrule
        \rowcolor{gray!20}
        Average Scores & 19.5 & \textbf{43.1} & 12.7 & \textbf{34.7} \\
        Gain    & ---  & {\color{teal}\textbf{+23.6}} & ---  & {\color{teal}\textbf{+22.0}} \\
        \bottomrule
    \end{tabular}%
    }
    \caption{\textbf{Real-World Results.}}
    \label{table:real-world-results}
\end{table}

\subsection{Ablation Study}
\label{subsec:Ablation Study}

\noindent\textbf{Impact of shared-mapping for dual codebooks.} To investigate the performance impact of the shared codebook mapping module on the RoboTwin benchmark\cite{Mu_2025_CVPR}, we conduct an ablation study using entirely independent codebooks. In this setting, direct training does not enforce any relationship between the codebooks. As shown in Table \ref{tab:combined_ablation}, removing the shared mapping and utilizing independent codebooks leads to a significant performance degradation. This result indicates the latent space constraints imposed by shared dual-codebook mapping are beneficial for coordination.

\noindent\textbf{Impact of two-stage training.} To validate the role of our two-stage training, we directly evaluate the end-to-end trained model on the RoboTwin benchmark\cite{Mu_2025_CVPR}. As shown in the table \ref{tab:combined_ablation}, the average success rate of the end-to-end trained model is significantly lower than that of the full two-stage model. This indicates that independent dual-arm pre-training followed by collaborative post-training plays a pivotal role in achieving effective coordinated control.

\begin{table}[ht]
\centering
\renewcommand{\arraystretch}{1.3}
\resizebox{\linewidth}{!}{
\begin{tabular}{cc|cc} 
\hline
\textbf{Shared Mapping} & \textbf{Two-stage training} & \textbf{Mean SR (\%)} & \textbf{Gain} \\ \hline
                        &                                & 32.6                  & -             \\ 
\checkmark              &                                & 42.1                  & \color{teal}{+9.5} \\ 
                        & \checkmark                     & 40.1                  & \color{teal}{+7.5} \\ 
\checkmark              & \checkmark                     & \textbf{51.8}         & \color{teal}{\textbf{+19.2}} \\ \hline
\end{tabular}}
\caption{\textbf{Impact of shared mapping and two-stage training}. Mean SR (\%) denotes the average success rate across all tasks, and Gain denotes the improvement in success rate relative to the first setting.}
\label{tab:combined_ablation}
\end{table}

\noindent\textbf{Impact of latent tokens and codebook size.} To manage the perception coordination between the two arms, we introduce latent tokens via a unidirectional attention mechanism to serve as a bridge. We conduct an ablation study comparing the effects of not using latent tokens, varying the number of latent tokens ($N$), and adjusting the codebook size ($K$). Since performance was similar in several cases, we select three representative parameter settings as examples. As shown in the table \ref{tab:latent_codebook_impact}, the model fails completely when latent tokens are omitted and the image and state features are quantized directly. Furthermore, when $N=8$ and $K=512$, the model's performance also degrades. This indicates that an appropriate number of latent tokens and a suitable corresponding codebook size are crucial for the effectiveness of the shared mapping constraint.

\begin{table}[t]
    \centering
    \small
    \setlength{\tabcolsep}{6pt}
    \renewcommand{\arraystretch}{1.1}
    \begin{tabular}{ccc}
        \toprule
        \textbf{Number of Latents} & \textbf{Codebook Size} & \textbf{Mean SR (\%)} \\
        \midrule
        0 & 256  & 0.0 \\
        4  & 256 & \textbf{51.8} \\
        8  & 512 & 40.2 \\
        \bottomrule
    \end{tabular}
    \caption{\textbf{Impact of latent tokens and codebook size.} Mean SR (\%) denotes the average success rate across all tasks.}
    \label{tab:latent_codebook_impact}
\end{table}

\section{Conclusion}
This work presented CUBic, a Coordinated and Unified framework for Bimanual perception and control, which bridges the long-standing divide between inter-arm independence and coordination.
By learning a shared tokenized representation across perception and control, CUBic enables coherent reasoning and synchronized dual-arm execution within a single unified policy.
Experiments show consistent performance gains over baselines, highlighting the effectiveness of unified modeling for coordinated manipulation.
This joint modeling proves crucial for capturing the complex spatio-temporal dynamics inherent in dual-arm tasks.
Our results validate that this unified policy structure surpasses traditional approaches.
This work takes a step toward scalable and generalizable visuomotor policies capable of reasoning jointly across multiple manipulators.

\clearpage

\section{Acknowledgements}
This work was supported by the New Generation Artificial Intelligence-National Science and Technology Major Project (2025ZD0122603). It was also supported by the Postdoctoral Fellowship Program and China Postdoctoral Science Foundation under Grant No. 2024M764093 and Grant No. BX20250485, the Beijing Natural Science Foundation under Grant No. 4254100, and by Beijing Advanced Innovation Center for Future Blockchain and Privacy Computing. It was also supported by the Brain Science and Brain-like Intelligence Technology — National Science and Technology Major Project (Grant No. 2022ZD0208800). It was also supported by the Young Elite Scientists Sponsorship Program of the Beijing High Innovation Plan (NO. 20250860).
{
    \small
    \bibliographystyle{ieeenat_fullname}
    \bibliography{main}
}

\end{document}